%% file: main.tex
\documentclass[lettersize,journal]{IEEEtran}
\usepackage{amsmath,amsfonts}
\usepackage{algorithmic}
\usepackage{algorithm}
\usepackage{array}
\usepackage[caption=false,font=normalsize,labelfont=sf,textfont=sf]{subfig}
\usepackage{textcomp}
\usepackage{stfloats}
\usepackage{url}
\usepackage{verbatim}
\usepackage{graphicx}
\usepackage{cite}
\usepackage{makecell}  
\usepackage{multirow}
\newcommand{\OurModel}[1]{CoSA}
\newcommand{\firstpara}[1]{\noindent\textbf{{#1}.}~~}

\usepackage{booktabs}
\usepackage{multirow}
\usepackage{graphicx}
\usepackage{booktabs}
\usepackage{makecell}
\usepackage[table]{xcolor}
\usepackage{pifont}
\usepackage{pgfplots}
\usepgfplotslibrary{groupplots}
\usetikzlibrary{patterns} 
\pgfplotsset{compat=1.18}
\usepackage{color}

\begin{document}


\title{Rethinking Transferable Adversarial Attacks on Point Clouds from a Compact Subspace  Perspective}

\author{Keke Tang, Xianheng Liu, Weilong Peng, Xiaofei Wang, Daizong Liu, Peican Zhu, Can Lu, and Zhihong Tian

\thanks{Keke Tang, Xianheng Liu, Can Lu and Zhihong Tian are with the Cyberspace Institute of Advanced Technology, Guangzhou University, Guangzhou, Guangdong 510006, China.}
\thanks{Weilong Peng is with the School of Computer Science and Cyber Engineering, Guangzhou University, Guangzhou, Guangdong 510006, China.}
\thanks{Xiaofei Wang is with the  Department of Automation, University of Science and Technology of China,
Hefei, Anhui 230052, China.}
\thanks{Daizong Liu is with the  Institute for Math \& AI, Wuhan University,
Wuhan, Hubei 430072, China.}
\thanks{Peican Zhu is with the School of Artificial Intelligence, Optics and
Electronics (iOPEN), Northwestern
Polytechnical University,
Xi’an, Shaanxi 710072, China.}
}


\markboth{Journal of \LaTeX\ Class Files,~Vol.~14, No.~8, August~2021}%
{Shell \MakeLowercase{\textit{et al.}}: A Sample Article Using IEEEtran.cls for IEEE Journals}


\maketitle

\input{0_abstract}
\input{1_intro}

\input{2_related_work}

\input{3_preliminaries}

\input{4_methodology}

\input{5_experiments}
\input{6_conclusion}

{
\bibliographystyle{IEEEtran}
\bibliography{ref}
}

\end{document}

%% file: 0_abstract.tex
\begin{abstract}

Transferable adversarial attacks on point clouds remain challenging, as existing methods often rely on model-specific gradients or heuristics that limit generalization to unseen architectures. 
In this paper, we rethink adversarial transferability from a compact subspace perspective and propose CoSA, a transferable attack framework that operates within a shared low-dimensional semantic space. 
Specifically, each point cloud is represented as a compact combination of class-specific prototypes that capture shared semantic structure, while adversarial perturbations are optimized within a low-rank subspace to induce coherent and architecture-agnostic variations. 
This design suppresses model-dependent noise and constrains perturbations to semantically meaningful directions, thereby improving cross-model transferability without relying on surrogate-specific artifacts. 
Extensive experiments on multiple datasets and network architectures demonstrate that CoSA consistently outperforms state-of-the-art transferable attacks, while maintaining competitive imperceptibility and robustness under common defense strategies. 
Codes will be made public upon paper acceptance.

\end{abstract}


%% file: 1_intro.tex
\section{Introduction}
\label{sec:intro}

\IEEEPARstart{R}{ecent} advances in deep learning have significantly improved 3D point cloud understanding, leading to remarkable performance in recognition, segmentation, and scene analysis~\cite{guo2020deep}.
Despite these achievements, DNN-based point cloud models remain sensitive to adversarial perturbations, where small and often imperceptible changes to input points can cause severe misclassifications~\cite{Xiang-2019-Generating}.
This vulnerability raises concerns in safety-critical applications such as autonomous driving and service robotics, where model reliability is crucial.
Understanding and mitigating these vulnerabilities through adversarial analysis is essential for building robust and trustworthy point cloud perception systems.

\begin{figure}[!t] 
\centering
\includegraphics[width=1\linewidth]{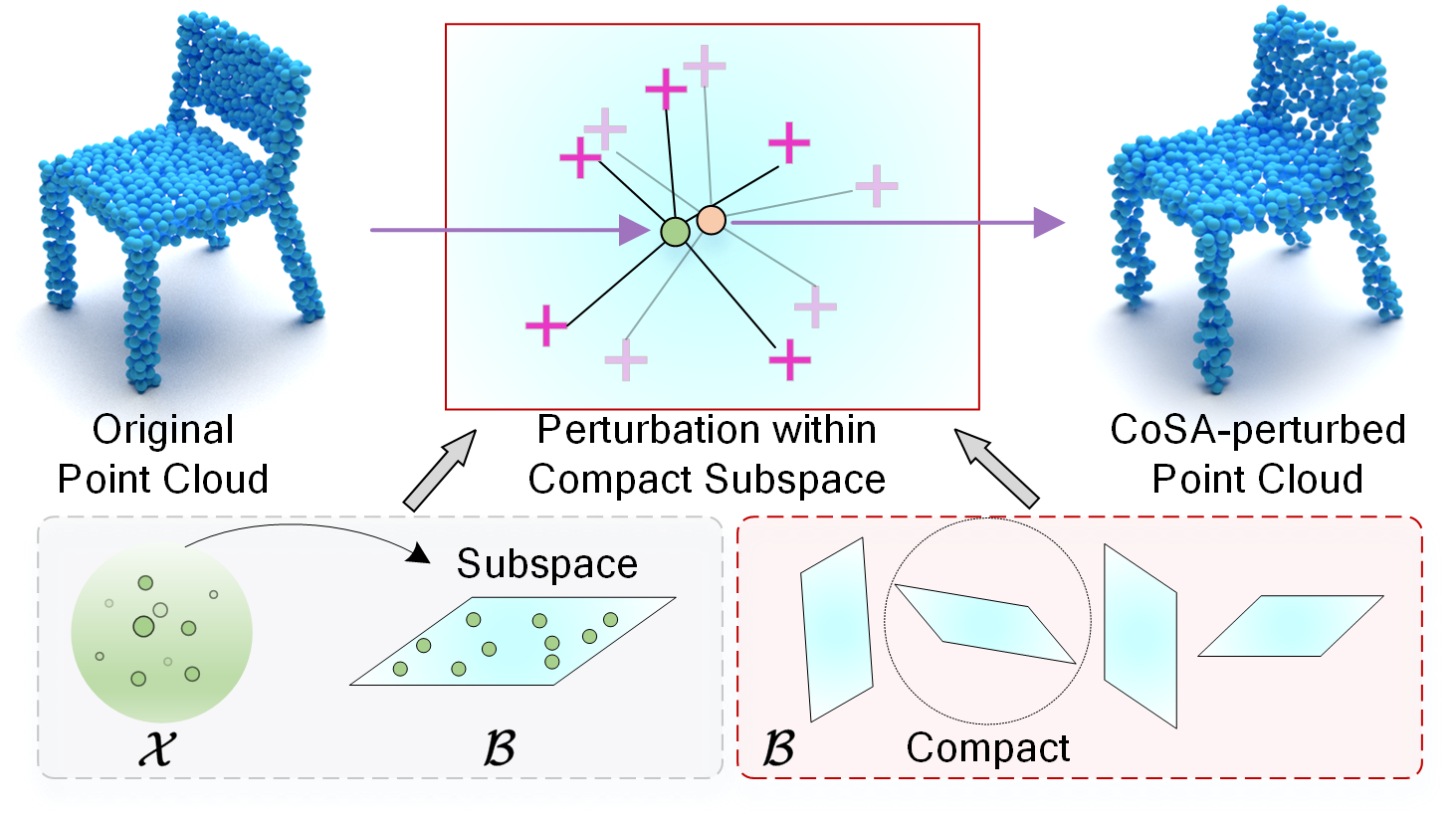}
\caption{
CoSA constrains adversarial perturbations from the ambient input space $\mathcal{X}$ to a compact subspace $\mathcal{B}$ that captures shared semantic representations,
thereby achieving stronger and more consistent transferability across models.
}

\label{fig:teaser} 
\end{figure}

Most existing point cloud attack methods are developed under white-box settings, where model gradients are fully accessible. These methods typically ensure imperceptibility by constraining geometric distortions or enforcing structural priors such as symmetry, uniformity, and surface consistency~\cite{wen-2020-geometry(geo),Tang-2024-FLAT,liu-2022-imperceptible(ITA),Huang-2022-shape(SI-Adv)}. Although effective on the source model, their reliance on model-specific gradients limits generalization to unseen models.
Recent work has explored augmentation-based approaches that improve robustness to geometric transformations~\cite{liu-2022-imperceptible(ITA)}, factorization-based approaches that decompose perturbations into complementary components~\cite{He-2023-PF,ANF}, and representation-based approaches that operate in latent or spectral spaces to capture model-invariant semantics~\cite{hamdi2020advpc,liu2022boosting,lee-2020-Shapeadv,Pang2025-CFG}. However, these approaches remain confined to model-specific feature spaces, often resulting in inconsistent semantic variations across architectures.
This motivates us to reinterpret point cloud attacks from a compact subspace perspective that promotes coherent, model-agnostic perturbations.

In representation learning, subspace learning has long been known to enhance generalization by capturing compact, low-dimensional representations that remain consistent across domains and models~\cite{zhuang2020comprehensive,pan2008transfer,li2015robust,shao2014generalized}. Constraining optimization within such subspaces encourages models to focus on shared semantic regularities rather than task-specific noise.
Following this principle, we hypothesize that the transferability of adversarial examples across point cloud models also benefits from compact subspace representations: perturbations confined to a shared subspace are more likely to manipulate semantic structures intrinsic to the data manifold and shared across architectures, thereby improving cross-model transferability.

Guided by this insight, we propose CoSA (Compact Subspace Attack), a transferable attack framework that models and perturbs compact subspace representations of point clouds (see Fig.~\ref{fig:teaser}).
Specifically, each point cloud is encoded into a latent space and represented as a sparse combination of class-specific prototypes, forming a compact base subspace that captures shared semantic structures.
Adversarial perturbations are further constrained within a shared low-rank subspace of prototype representations, promoting coherent variations along model-agnostic directions.
By integrating semantic and perturbation subspaces within a unified compact framework, CoSA produces consistent and transferable adversarial examples.
Comprehensive evaluations across multiple source–target model pairs and datasets demonstrate that CoSA improves transferability and outperforms state-of-the-art attack methods. We also show that CoSA remains effective under defense settings.

Overall, our contribution is summarized as follows:
\begin{itemize}
\item We introduce a novel compact subspace  perspective to interpret and enhance the transferability of adversarial attacks on DNN-based point cloud models.
\item We propose CoSA, a compact subspace–based adversarial attack framework that encodes each point cloud as a sparse combination of class-specific prototypes and constrains perturbations within a shared low-rank subspace.
\item We show by experiments that CoSA achieves strong transferability across point cloud DNN architectures and datasets, outperforming state-of-the-art attack methods.
\end{itemize}

%% file: 2_related_work.tex
\section{Related Work}

\subsection{Adversarial Attacks on Point Clouds}
Adversarial attacks on point clouds have been extensively studied in recent years.
Due to the unordered and irregular nature of point clouds, most existing works focus on designing perturbations that preserve geometric structures while remaining adversarial.
According to the attack mechanism, prior methods can be broadly categorized into addition-based, deletion-based, and perturbation-based attacks.

Addition-based attacks insert carefully crafted points into the original point cloud to mislead the classifier~\cite{Xiang-2019-Generating}.
Deletion-based attacks remove critical or salient points identified through saliency or gradient information~\cite{Zheng-2019-PCDSaliency,Wicker-2019-IterSaliencyOcc}.
Perturbation-based attacks directly modify point coordinates under geometric constraints and have received the most attention due to their flexibility and effectiveness~\cite{zhao2020isometry,li2025nopain,lou2024hide,shi2022shape}.

Early perturbation-based methods mainly adapted 2D adversarial techniques, such as FGSM and C\&W, to 3D classifiers~\cite{Xiang-2019-Generating,liu-2019-extending}.
Subsequent works incorporated geometry-aware constraints, including curvature preservation, symmetry enforcement, uniformity regularization, and surface consistency, to improve imperceptibility~\cite{wen-2020-geometry(geo),tang2024symattack,Tang-2024-FLAT,liu-2022-imperceptible(ITA),Huang-2022-shape(SI-Adv),zhang2024curvature}.
While these methods achieve strong white-box attack performance, their perturbations are often tightly coupled to the surrogate model gradients, leading to limited generalization across unseen architectures.

\subsection{Transferable Attacks on Point Clouds}
Transferable adversarial attacks aim to generate perturbations that remain effective across different models, which is essential for evaluating security risks in realistic black-box settings.
Although early studies primarily focused on white-box optimization, improving transferability has recently become a central research topic.

Existing transferable attack methods can be roughly divided into three categories.
Augmentation-based methods improve transferability by enforcing robustness to geometric transformations during optimization~\cite{liu-2022-imperceptible(ITA)}.
Factorization-based methods enhance generalization by decomposing perturbations into complementary components.
He et al.~\cite{He-2023-PF} proposed randomly factorizing perturbations into multiple sub-perturbations to encourage cross-model consistency, while Chen et al.~\cite{ANF} introduced feature-space noise factorization to reduce reliance on surrogate gradients.
Representation-based approaches operate in latent or spectral domains to capture model-invariant semantics.
Hamdi et al.~\cite{hamdi2020advpc} employed a denoising autoencoder to enforce adversariality on both original and reconstructed point clouds, while Liu et al.~\cite{liu2022boosting} perturbed low-frequency graph Fourier components to promote smoother and more transferable distortions.
Later works, such as ShapeAdv~\cite{lee-2020-Shapeadv} and CFG~\cite{Pang2025-CFG}, further explored latent-space perturbations and feature-guided constraints to improve transferability.

Despite these advances, existing representation-based methods typically operate in model-dependent feature spaces and do not explicitly constrain perturbations to lie in a shared semantic structure across architectures.
In contrast, our method introduces a compact subspace perspective that explicitly models shared semantic representations and restricts adversarial perturbations within a low-dimensional, architecture-agnostic subspace, leading to improved transferability.

\subsection{DNN-based Point Cloud Classification}
Deep learning has significantly advanced point cloud classification.
Early approaches voxelized point clouds and applied 3D convolutional networks~\cite{Maturana-2015-voxnet}, but suffered from high memory consumption and limited spatial resolution.
PointNet~\cite{Qi-2017-Pointnet} and PointNet++~\cite{Qi-2017-Pointnet++} enabled direct learning from raw point sets, laying the foundation for later developments.

Subsequent works explored point-wise convolution~\cite{Wu-2019-Pointconv,Li-2018-PointCNN}, graph-based aggregation~\cite{Wang-2019-DGCNN,Zhao-2019-Pointweb}, and more recently Transformer- and Mamba-based architectures~\cite{zhao2021PT,wu2024pT3,liang2024pointmamba,han2024mamba3d}.
These diverse architectures exhibit substantially different inductive biases and internal representations, making transferable attacks particularly challenging.
This work investigates adversarial transferability across such representative classifiers.

\subsection{Compact Subspace Representation}
Compact subspace learning aims to represent high-dimensional data using low-dimensional structures that preserve intrinsic semantics and geometric relationships~\cite{li2015robust,7053941}.
By introducing low-rank or sparse constraints, compact subspace models extract shared and stable representations and have been widely applied in domain adaptation and transfer learning~\cite{pan2008transfer,zhuang2020comprehensive,yang2022marginal}.

In this work, compact subspace representation is leveraged from an adversarial perspective.
Instead of directly optimizing perturbations in the high-dimensional input space or model-specific latent spaces, we construct a shared compact subspace that captures class-level semantic structure and restrict adversarial perturbations to a low-rank variation space.
This design suppresses model-specific noise and promotes perturbations that generalize across architectures, providing a principled foundation for transferable adversarial attacks on point clouds.

%% file: 3_preliminaries.tex
\section{Problem Formulation}

\subsection{Preliminaries on Point Cloud Attacks}
Let \(P\in\mathbb{R}^{n\times 3}\) denote a point cloud with ground-truth label \(y\in\{1,\dots,Z\}\),
where \(Z\) is the total number of classes.
In a transferable attack setting, the adversary has access only to a surrogate classifier
\(f_s:\mathbb{R}^{n\times 3}\to\{1,\dots,Z\}\)
and aims to generate an adversarial example that also fools an unseen target classifier \(f_t\)
trained on the same label space.
Following standard practice, the (untargeted) perturbation \(\delta\) is commonly obtained by solving
\begin{equation}
\begin{aligned}
\delta^{\star}
= \arg\min_{\delta}\;&
\bigl[L_{\mathrm{mis}}(f_s,P+\delta,y)
+\lambda_{\mathrm{per}}L_{\mathrm{per}}(P,P+\delta)\bigr],\\
\text{s.t.}\;& D(P,P+\delta)\le\varepsilon,
\end{aligned}
\label{eq:surrogate_opt}
\end{equation}
where \(L_{\mathrm{mis}}\) promotes misclassification 
(e.g., the negation of the cross-entropy loss), 
\(L_{\mathrm{per}}\) penalizes geometric distortion 
(e.g., geometry-aware distances such as Chamfer or Hausdorff metrics), 
and \(D(\cdot,\cdot)\le\varepsilon\) enforces an explicit geometric bound 
(typically an \(\ell_p\) norm such as \(\ell_\infty\)). 
The scalar \(\lambda_{\mathrm{per}}\ge 0\) balances attack strength and imperceptibility.
The optimal perturbation is denoted as \(\delta^{\star}\), and the corresponding adversarial point cloud is
\begin{equation}
P^{\mathrm{adv}} = P + \delta^{\star}.
\end{equation}
The transferable attack succeeds  if \(f_t(P^{\mathrm{adv}})\neq y\).

\firstpara{Discussion}
Direct input-space optimization usually produces high-dimensional, instance-specific perturbations that heavily rely on model gradients and overfit surrogate models, resulting in poor cross-model transferability.

Compact subspace learning has been shown to enhance generalization and transferability by capturing shared low-dimensional representations across domains~\cite{zhuang2020comprehensive,pan2008transfer}.
Inspired by this principle, we hypothesize that constraining adversarial optimization within a \emph{compact subspace} encourages perturbations toward model-agnostic and semantically coherent variations, improving cross-model transferability.

\subsection{Point Cloud Attacks in a Compact Subspace}
\label{sec:formulation}

We posit that transferable adversarial examples can be modeled within a compact shared structure that separates stable object representations from transferable variations.  
Let the ambient input space \(\mathcal{X}\subset\mathbb{R}^{n\times3}\) denote the set of all point clouds, and assume the existence of a shared low-dimensional subspace
\begin{equation}
\mathcal{S}_{\mathrm{shared}}=\mathcal{B}\oplus\mathcal{S}\subseteq\mathcal{X},
\label{eq:shared_structure_compact}
\end{equation}
where
\begin{itemize}
  \item \(\mathcal{B}\) (\emph{base subspace}) is a compact subspace that preserves intrinsic semantic representations of objects, enabling faithful reconstruction of typical instances with only a small number of basis elements; and
  \item \(\mathcal{S}\) (\emph{perturbation subspace}) is a compact, low-dimensional subspace that captures coherent and transferable adversarial variations across instances and models, allowing perturbations to evolve along shared semantic directions.
\end{itemize}

Both subspaces are compact in nature, providing low-dimensional representations that emphasize shared and semantically meaningful components.  
Consequently, \(\dim(\mathcal{B})\ll\dim(\mathcal{X})\) and \(\dim(\mathcal{S})\ll\dim(\mathcal{B})\).  
We assume that the mappings \(\Pi_{\mathcal B}:\mathcal{X}\to\mathcal{B}\) and \(\Pi_{\mathcal S}:\mathcal{B}\to\mathcal{S}\) are approximately invertible within their respective domains, with inverse mappings \(\Pi_{\mathcal B}^{-1}\) and \(\Pi_{\mathcal S}^{-1}\) representing reconstruction operators.

Adversarial optimization is thus restricted to perturbations 
\(\delta_{\mathcal B}\in\mathcal{B}\)
that are projected through \(\mathcal{S}\) before reconstruction:
\begin{equation}
\begin{aligned}
\delta_{\mathcal B}^{\star}
=\arg\min_{\delta_{\mathcal B}\in\mathcal{B}}\;&
\Big[L_{\mathrm{mis}}\!\big(f_s,P',y\big)
+\lambda_{\mathrm{per}}L_{\mathrm{per}}(P,P')\Big],\\
\text{s.t.}\quad &
P'=\Pi_{\mathcal B}^{-1}\!\big(\Pi_{\mathcal B}(P)
+\Pi_{\mathcal S}^{-1}(\Pi_{\mathcal S}(\delta_{\mathcal B}))\big),\\
& D(P',P)\le\varepsilon.
\end{aligned}
\label{eq:compact_subspace_attack}
\end{equation}
The corresponding adversarial example is reconstructed as
\begin{equation}
P^{\mathrm{adv}}
=\Pi_{\mathcal B}^{-1}\!\Big(
\Pi_{\mathcal B}(P)+\Pi_{\mathcal S}^{-1}\!\big(
\Pi_{\mathcal S}(\delta_{\mathcal B}^{\star})
\big)\Big).
\end{equation}

By confining adversarial optimization to compact subspaces that emphasize shared, semantically meaningful modes and suppress model-specific noise, the resulting \(P^{\mathrm{adv}}\) 
naturally favors perturbations that generalize across models rather than overfitting the surrogate.


%% file: 4_methodology.tex
\begin{figure*}[!t]
\centering
\includegraphics[width=0.77\linewidth]{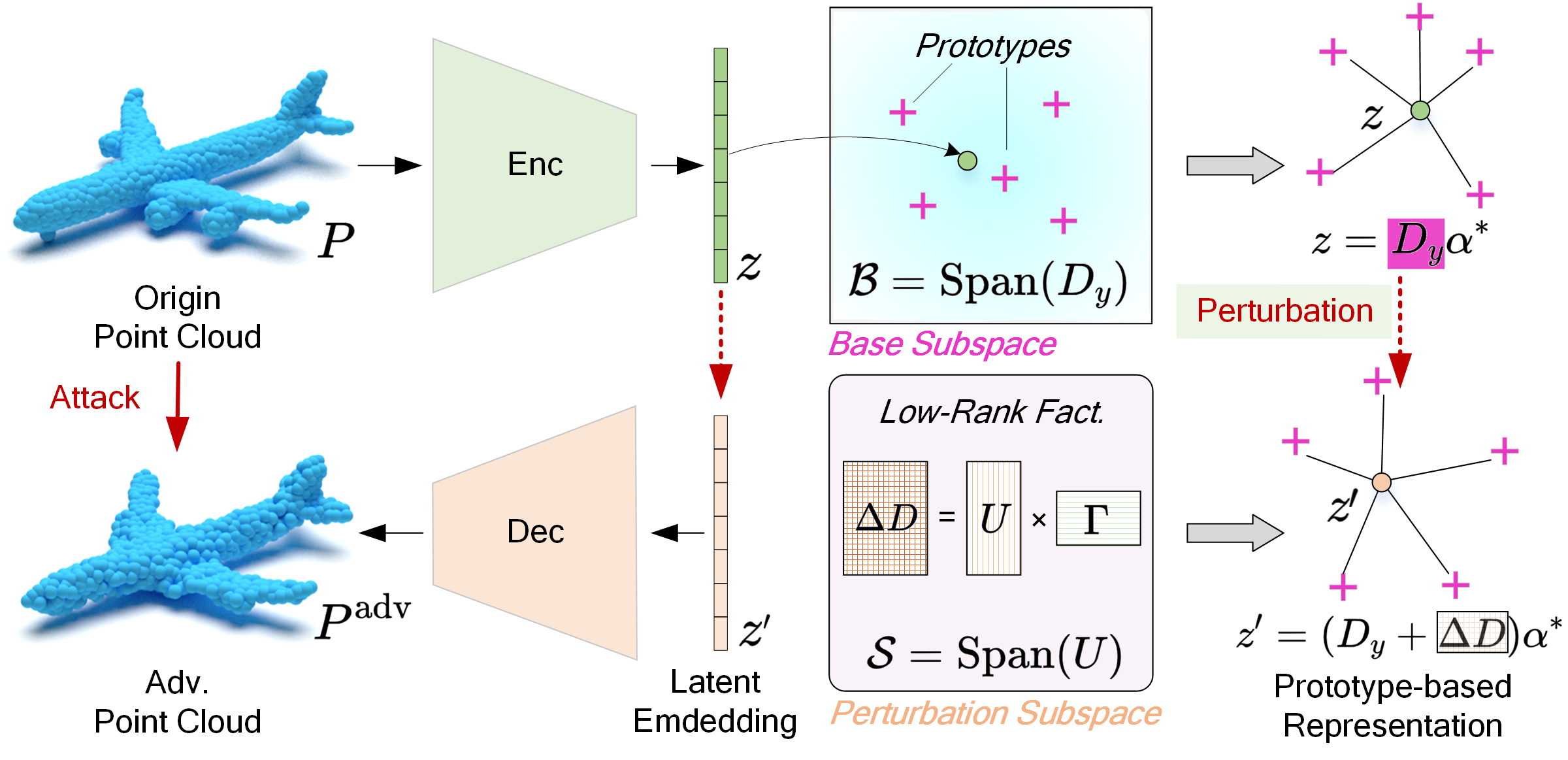} 
\caption{
Overview of the proposed CoSA framework.
The encoder embeds the input point cloud into a prototype-guided base subspace  \(\mathcal{B}\) capturing class-level semantics, while a low-rank perturbation subspace  \(\mathcal{S}\) models coherent variations.
Joint optimization yields a perturbed latent representation decoded into an adversarial point cloud with improved transferability.
}
\label{fig:framework}
\end{figure*}

\section{Method}
\label{sec:method}

In this section, we present CoSA (Compact Subspace Attack), a transferable adversarial attack framework instantiated from the compact-subspace formulation in Sec.~\ref{sec:formulation}.
CoSA decomposes each point cloud into two latent subspaces:
a prototype-guided base subspace preserving class semantics, and a low-rank perturbation subspace capturing coherent and transferable variations (see Fig.~\ref{fig:framework}).
This compact parameterization suppresses instance- and model-specific noise while maintaining semantic structure, thereby enhancing cross-model transferability.

\subsection{Constructing Prototype-Guided Base Subspace}

We instantiate the base subspace \(\mathcal{B}\) in a compact latent domain obtained from a pretrained autoencoder (AE):
\begin{equation}
z = \mathrm{Enc}(P), \qquad \hat{P} = \mathrm{Dec}(z),
\end{equation}
where \(\mathrm{Enc}:\mathbb{R}^{n\times3}\!\to\!\mathbb{R}^d\) and \(\mathrm{Dec}:\mathbb{R}^d\!\to\!\mathbb{R}^{n\times3}\).
The autoencoder is pretrained for faithful reconstruction and kept fixed during subsequent subspace construction.

To capture shared semantics, we construct a class-wise prototype dictionary \(D_y \in \mathbb{R}^{d\times m_y}\) 
by applying k-means clustering to latent embeddings of class \(y\), where the \(m_y\) cluster centers serve as class-specific prototypes. The prototype span \(\mathcal{B} = \mathrm{span}(D_y)\) thus defines the latent base subspace that preserves class-level semantic structures.

Given a latent embedding \(z\), its representation is obtained by sparse reconstruction over the corresponding class dictionary:
\begin{equation}
\alpha^{\star} = \arg\min_{\alpha} \|z - D_y \alpha\|_2^2 + \lambda_{\mathrm{spa}} \|\alpha\|_1.
\end{equation}

\subsection{Modeling Low-Rank Perturbation Subspace}

Building on the base subspace \(\mathcal{B}\), we construct a compact \emph{low-rank perturbation subspace} \(\mathcal{S}\) 
to capture coherent and transferable adversarial variations.  
The perturbation matrix is parameterized as
\begin{equation}
\Delta D_y = U\Gamma,
\end{equation}
where \(U\in\mathbb{R}^{d\times r}\) is an orthonormal basis spanning the perturbation subspace,
i.e., \(\mathcal{S} = \mathrm{span}(U)\), 
and \(\Gamma\in\mathbb{R}^{r\times m_y}\) controls the perturbation magnitudes associated with the \(m_y\) prototype bases of class \(y\).  
The perturbed latent embedding is then written as
\begin{equation}
z' = (D_y + \Delta D_y)\alpha^\star = (D_y + U\Gamma)\alpha^\star.
\end{equation}
This low-rank factorization enforces correlation among prototype perturbations 
and constrains them to vary along a few shared, model-agnostic directions in the latent space, 
thereby forming a compact and structured representation of transferable variations.

\subsection{Compact Subspace Attack Objective}

Integrating the base subspace \(\mathcal{B}\) and the perturbation subspace \(\mathcal{S}\), 
CoSA jointly optimizes the perturbation parameters \((U,\Gamma)\) under a bounded reconstruction constraint.  
The objective is formulated as
\begin{equation}
\begin{aligned}
\min_{U,\Gamma}\quad &
L_{\mathrm{mis}}\big(f_s,\mathrm{Dec}(z'),y\big)
+\lambda_{\mathrm{per}}L_{\mathrm{per}}\big(P,\mathrm{Dec}(z')\big)\\
&\quad+\lambda_{\mathrm{rank}}\|\Gamma\|_{*}
+\lambda_{\mathrm{ort}}\big\|U^\top U - I\big\|_F^2,\\
 & \text{s.t.}\quad 
z' = (D_y + U\Gamma)\alpha^\star,\\
&\quad\quad D\!\big(\mathrm{Dec}(z'),P\big) \le \varepsilon.
\end{aligned}
\label{eq:final_obj_method}
\end{equation}
where the nuclear norm term \(\|\Gamma\|_*\) encourages a low-rank structure within the perturbation subspace, 
promoting correlated and compact variations across prototypes, 
while the orthogonality term \(\|U^\top U - I\|_F^2\) stabilizes the basis directions of \(U\) 
to maintain orthonormal perturbation modes.  
\(\lambda_{\mathrm{rank}}\) and \(\lambda_{\mathrm{ort}}\) are hyperparameters 
that balance attack strength, compactness, and basis stability.

After obtaining the optimal parameters \((U^\star,\Gamma^\star)\),  
the adversarial point cloud is reconstructed by decoding the perturbed latent embedding:
\begin{equation}
P^{\mathrm{adv}} = \mathrm{Dec}\!\big ((D_y+U^\star\Gamma^\star)\alpha^\star\big).
\label{eq:dec_attack}
\end{equation}
This decoding step maps compact latent perturbations back to the 3D input domain,  
preserving semantic coherence and producing transferable adversarial point clouds.

%% file: 5_experiments.tex
\section{Experiments}
\label{sec:exp}


\begin{table*}[!t]
\centering
\caption{Transferability performance of different attack methods on ModelNet40 and ScanObjectNN.  
Transferability is evaluated by the attack success rate (ASR, \%) on target models using adversarial examples generated from source models, under $\ell_\infty$ perturbation budgets of $\epsilon\!=\!0.18$ and $0.45$.  
Bold indicates the best transferability, while gray denotes white-box results.}
\label{tab:transfer_table_all}

\setlength{\tabcolsep}{0.87mm}{
\scalebox{1}{
\begin{tabular}{@{}c l|cccc|cccc|cccc|cccc@{}}
\toprule
\multirow{4}{*}{\makecell{Victim\\Model}} & \multirow{4}{*}{\makecell{Attack\\Method}} &
\multicolumn{8}{c|}{\textbf{ModelNet40}} &
\multicolumn{8}{c}{\textbf{ScanObjectNN}}\\
\cmidrule(lr){3-10}\cmidrule(lr){11-18}
& & \multicolumn{4}{c|}{$\epsilon=0.18$} & \multicolumn{4}{c|}{$\epsilon=0.45$} & \multicolumn{4}{c|}{$\epsilon=0.18$} & \multicolumn{4}{c}{$\epsilon=0.45$} \\

\cmidrule(lr){3-6}\cmidrule(lr){7-10}\cmidrule(lr){11-14}\cmidrule(lr){15-18}
& & \scalebox{0.90}{PointNet} & \scalebox{0.90}{DGCNN} & \scalebox{0.90}{PCT} & \scalebox{0.90}{\makecell{Point-\\Mamba}}
  & \scalebox{0.90}{PointNet} & \scalebox{0.90}{DGCNN} & \scalebox{0.90}{PCT} & \scalebox{0.90}{\makecell{Point-\\Mamba}}
  & \scalebox{0.90}{PointNet} & \scalebox{0.90}{DGCNN} & \scalebox{0.90}{PCT} & \scalebox{0.90}{\makecell{Point-\\Mamba}}
  & \scalebox{0.90}{PointNet} & \scalebox{0.90}{DGCNN} & \scalebox{0.90}{PCT} & \scalebox{0.90}{\makecell{Point-\\Mamba}} \\
\midrule

\multirow{9}{*}{\rotatebox[origin=c]{90}{PointNet}}
& 3D-Adv     & \textcolor{gray}{100.0} & 1.4  & 4.9  & 1.3  & \textcolor{gray}{100.0} & 1.5  & 5.2  & 1.0     
             & \textcolor{gray}{100.0} & 9.0  & 6.0  & 1.2  & \textcolor{gray}{100.0} & 9.0  & 5.5  & 1.6 \\
& KNN        & \textcolor{gray}{99.7}  & 6.0  & 6.9  & 5.6  & \textcolor{gray}{99.7}  & 6.4  & 6.9  & 5.7     
             & \textcolor{gray}{100.0} & 49.3 & 54.4 & 45.2 & \textcolor{gray}{100.0} & 49.8 & 54.0 & 45.1 \\
& AdvPC      & \textcolor{gray}{100.0} & 22.4 & 16.3 & 29.0 & \textcolor{gray}{100.0} & 22.6 & 16.2 & 28.9    
             & \textcolor{gray}{100.0} & 36.4 & 49.8 & 35.2 & \textcolor{gray}{100.0} & 36.4 & 48.3 & 35.2 \\
& AOF        & \textcolor{gray}{100.0} & 29.4 & 25.7 & 36.5 & \textcolor{gray}{100.0} & 30.9 & 26.9 & 36.7    
             & \textcolor{gray}{100.0} & 44.8 & 53.3 & 43.4 & \textcolor{gray}{100.0} & 44.4 & 52.3 & 43.6 \\
& PF-Attack  & \textcolor{gray}{94.0}  & 24.8 & 23.4 & 38.2 & \textcolor{gray}{94.0}  & 28.8 & 28.7 & 48.5    
             & \textcolor{gray}{92.9}  & 40.3 & 45.6 & 47.1 & \textcolor{gray}{93.3}  & 40.8 & 47.1 & 48.5 \\
& ShapeAdv   & \textcolor{gray}{100.0} & 23.5 & 21.7 & 19.1 & \textcolor{gray}{100.0} & 26.2 & 22.1 & 19.3  
             & \textcolor{gray}{100.0} & 20.4 & 21.3 & 16.2 & \textcolor{gray}{100.0} & 23.1 & 24.6 & 18.4 \\
& MAT-Adv        & \textcolor{gray}{91.7}  & 30.2 & 28.7 & 35.2 & \textcolor{gray}{95.8}  & 36.8 & 35.3 & 37.8    
             & \textcolor{gray}{99.8}  & 44.9 & 54.2 & 42.3 & \textcolor{gray}{99.7}  & 46.9 & 50.1 & 43.1 \\
& CFG        & \textcolor{gray}{90.4}  & 39.3 & 30.4 & 50.8 & \textcolor{gray}{92.0}  & 49.3 & 48.6 & 56.8    
             & \textcolor{gray}{89.9}  & 51.5 & 62.8 & 52.8 & \textcolor{gray}{90.4}  & 54.3 & 63.8 & 56.1 \\
& Ours       & \textcolor{gray}{97.7}  & \textbf{55.1} & \textbf{48.8} & \textbf{62.7} & \textcolor{gray}{100.0} 
             & \textbf{58.1} & \textbf{54.2} & \textbf{65.6}    
             & \textcolor{gray}{97.7}  & \textbf{66.2} & \textbf{73.6} & \textbf{63.6} & \textcolor{gray}{99.7}  
             & \textbf{69.2} & \textbf{75.7} & \textbf{67.2} \\
\midrule

\multirow{9}{*}{\rotatebox[origin=c]{90}{DGCNN}}
& 3D-Adv     & 3.6  & \textcolor{gray}{100.0} & 5.7  & 7.5  & 2.9  & \textcolor{gray}{100.0} & 6.8  & 7.4     
             & 4.3  & \textcolor{gray}{100.0} & 29.8 & 21.1 & 4.3  & \textcolor{gray}{100.0} & 29.1 & 21.8 \\
& KNN        & 7.2  & \textcolor{gray}{100.0} & 8.6  & 34.3 & 6.8  & \textcolor{gray}{99.6}  & 9.3  & 34.5    
             & 40.9 & \textcolor{gray}{100.0} & 68.2 & 62.1 & 40.7 & \textcolor{gray}{100.0} & 66.3 & 61.3 \\
& AdvPC      & 19.6 & \textcolor{gray}{95.7}  & 18.2 & 48.8 & 21.3 & \textcolor{gray}{100.0} & 18.4 & 47.6    
             & 18.5 & \textcolor{gray}{92.9}  & 61.6 & 52.5 & 18.5 & \textcolor{gray}{93.0}  & 60.9 & 52.3 \\
& AOF        & 20.7 & \textcolor{gray}{100.0} & 29.7 & 63.3 & 24.3 & \textcolor{gray}{100.0} & 34.1 & 63.7    
             & 17.5 & \textcolor{gray}{100.0} & 58.1 & 48.0 & 18.1 & \textcolor{gray}{100.0} & 57.9 & 48.1 \\
& PF-Attack  & 21.3 & \textcolor{gray}{91.6}  & 25.7 & 65.2 & 26.1 & \textcolor{gray}{91.6}  & 32.7 & 66.1    
             & 52.8 & \textcolor{gray}{91.1}  & 65.3 & 57.7 & 54.6 & \textcolor{gray}{91.2}  & 63.9 & 58.0 \\
& ShapeAdv   & 23.8 & \textcolor{gray}{100.0} & 22.9 & 17.4 & 24.1 & \textcolor{gray}{99.8}  & 25.0 & 20.2 
             & 19.6 & \textcolor{gray}{100.0} & 18.7 & 14.2 & 22.4 & \textcolor{gray}{99.9}  & 20.9 & 16.1 \\
& MAT-Adv        & 27.8 & \textcolor{gray}{95.1}  & 21.8 & 53.9 & 37.6 & \textcolor{gray}{99.1}  & 29.5 & 55.4    
             & 50.3 & \textcolor{gray}{100.0} & 62.9 & 55.2 & 52.4 & \textcolor{gray}{100.0} & 59.5 & 53.2 \\
& CFG        & 29.3 & \textcolor{gray}{94.4}  & 36.5 & 70.9 & 39.0 & \textcolor{gray}{90.0}  & 48.5 & 72.5    
             & 56.7 & \textcolor{gray}{89.1}  & 70.6 & 62.9 & 58.2 & \textcolor{gray}{89.1}  & 70.2 & 63.7 \\
& Ours       & \textbf{39.7} & \textcolor{gray}{97.6}  & \textbf{45.7} & \textbf{73.4} & \textbf{48.5} 
             & \textcolor{gray}{99.0}  & \textbf{56.2} & \textbf{75.9}    
             & \textbf{64.1} & \textcolor{gray}{99.4}  & \textbf{77.7} & \textbf{67.9} & \textbf{65.7} & \textcolor{gray}{99.7}  & \textbf{76.6} & \textbf{71.6} \\
\midrule

\multirow{9}{*}{\rotatebox[origin=c]{90}{PCT}}
& 3D-Adv     & 1.3  & 2.2  & \textcolor{gray}{100.0} & 1.1  & 1.2  & 2.3  & \textcolor{gray}{100.0} & 1.2     
             & 3.6  & 18.3 & \textcolor{gray}{100.0} & 20.7 & 4.0  & 18.5 & \textcolor{gray}{100.0} & 20.3 \\
& KNN        & 16.6 & 31.6 & \textcolor{gray}{99.9 } & 45.7 & 25.9 & 34.8 & \textcolor{gray}{99.9 } & 47.1    
             & 11.6 & 60.6 & \textcolor{gray}{100.0} & 47.9 & 10.5 & 61.3 & \textcolor{gray}{100.0} & 46.3 \\
& AdvPC      & 5.7  & 18.3 & \textcolor{gray}{100.0} & 41.7 & 7.2  & 20.1 & \textcolor{gray}{100.0} & 42.3    
             & 12.5 & 44.5 & \textcolor{gray}{98.5 } & 45.1 & 12.6 & 44.4 & \textcolor{gray}{98.5 } & 44.7 \\
& AOF        & 9.8  & 24.5 & \textcolor{gray}{100.0} & 48.0 & 13.4 & 27.7 & \textcolor{gray}{100.0} & 48.2    
             & 13.4 & 44.1 & \textcolor{gray}{95.6 } & 43.5 & 13.7 & 44.3 & \textcolor{gray}{95.7 } & 43.4 \\
& PF-Attack  & 19.2 & 32.9 & \textcolor{gray}{93.6 } & 68.7 & 23.9 & 37.7 & \textcolor{gray}{95.6 } & 69.9    
             & 50.5 & 51.1 & \textcolor{gray}{94.3 } & 53.6 & 52.6 & 49.7 & \textcolor{gray}{94.3 } & 54.3 \\
& ShapeAdv   & 24.9 & 23.2 & \textcolor{gray}{96.8 } & 18.3 & 27.3 & 25.9 & \textcolor{gray}{98.2 } & 20.4 
             & 20.5 & 19.3 & \textcolor{gray}{99.2 } & 14.7 & 23.6 & 22.2 & \textcolor{gray}{100.0} & 16.8 \\
& MAT-Adv        & 23.3 & 27.9 & \textcolor{gray}{94.3 } & 50.2 & 25.1 & 30.0 & \textcolor{gray}{94.6 } & 51.3    
             & 49.3 & 52.7 & \textcolor{gray}{96.8 } & 48.3 & 49.9 & 53.1 & \textcolor{gray}{96.5 } & 52.4 \\
& CFG        & 30.6 & 41.4 & \textcolor{gray}{88.8 } & \textbf{75.6} & 31.8 & 50.4 & \textcolor{gray}{90.4 } 
             & \textbf{74.8 }   
             & 51.8 & 61.9 & \textcolor{gray}{87.2 } & 61.1 & 54.1 & 63.2 & \textcolor{gray}{86.8 } & 61.8 \\
& Ours       & \textbf{35.5} & \textbf{58.6} & \textcolor{gray}{97.7} & 74.3 & \textbf{38.6} & \textbf{58.1} 
             & \textcolor{gray}{98.5 } & 72.9     
             & \textbf{54.3} & \textbf{72.2} & \textcolor{gray}{100.0 } & \textbf{72.3} & \textbf{54.5} & \textbf{75.3} & \textcolor{gray}{100.0 } & \textbf{72.9} \\
\midrule

\multirow{9}{*}{\rotatebox[origin=c]{90}{PointMamba}}
& 3D-Adv     & 0.9  & 1.8  & 4.0  & \textcolor{gray}{100.0} & 1.0  & 1.9  & 4.1  & \textcolor{gray}{100.0}    
             & 3.5  & 19.0 & 26.5 & \textcolor{gray}{100.0} & 3.0  & 19.1 & 26.7 & \textcolor{gray}{100.0} \\
& KNN        & 6.5  & 16.3 & 9.8  & \textcolor{gray}{99.9 } & 6.5  & 17.5 & 9.2  & \textcolor{gray}{99.9 }    
             & 12.8 & 65.1 & 69.1 & \textcolor{gray}{100.0} & 12.2 & 66.3 & 70.1 & \textcolor{gray}{100.0} \\
& AdvPC      & 8.0  & 28.0 & 13.6 & \textcolor{gray}{98.0 } & 8.1  & 27.6 & 12.7 & \textcolor{gray}{98.6 }    
             & 16.7 & 55.8 & 67.4 & \textcolor{gray}{99.1 } & 16.7 & 55.7 & 67.7 & \textcolor{gray}{99.1 }\\
& AOF        & 10.4 & 40.7 & 17.7 & \textcolor{gray}{100.0}& 11.8 & 42.5 & 19.9 & \textcolor{gray}{100.0 }   
             & 16.1 & 54.3 & 63.1 & \textcolor{gray}{100.0} & 16.0 & 55.1 & 62.8 & \textcolor{gray}{100.0}\\
& PF-Attack  & 14.0 & 18.4 & 17.2 & \textcolor{gray}{100.0} & 15.6 & 19.2 & 23.6 & \textcolor{gray}{100.0}    
             & 50.7 & 51.2 & 57.2 & \textcolor{gray}{96.5 } & 50.5 & 51.0 & 57.7 & \textcolor{gray}{96.4 }\\
& ShapeAdv   & 19.4 & 18.1 & 17.9 & \textcolor{gray}{94.2 } & 22.1 & 20.7 & 20.1 & \textcolor{gray}{96.7} 
             & 15.7 & 14.9 & 14.5 & \textcolor{gray}{100.0} & 18.9 & 17.5 & 17.1 & \textcolor{gray}{100.0} \\
& MAT-Adv        & 17.3 & 19.9 & 20.6 & \textcolor{gray}{93.9 } & 18.2 & 20.5 & 22.3 & \textcolor{gray}{94.8 }     
             & 44.3 & 46.2 & 53.8 & \textcolor{gray}{93.7 } & 44.9 & 47.3 & 55.2 & \textcolor{gray}{94.1 } \\
& CFG        & 10.4 & 12.5 & 10.5 & \textcolor{gray}{100.0} & 12.5 & 16.1 & 13.3 & \textcolor{gray}{100.0}    
             & 54.1 & 61.7 & 71.2 & \textcolor{gray}{84.4 } & 55.3 & 62.6 & 72.3 & \textcolor{gray}{84.3 }\\
& Ours       & \textbf{39.1} & \textbf{45.5} & \textbf{33.2} & \textcolor{gray}{96.2 } & \textbf{44.3} 
             & \textbf{48.7} & \textbf{39.3} & \textcolor{gray}{97.0 }     
             & \textbf{58.7} & \textbf{72.8} & \textbf{80.8} & \textcolor{gray}{100.0}  & \textbf{60.0} & \textbf{75.8}  & \textbf{81.3} & \textcolor{gray}{100.0} \\
\bottomrule
\end{tabular}
}}
\end{table*}

\subsection{Experimental Setup}

\firstpara{Implementation}
We implement CoSA using PyTorch. 
For the autoencoder backbone, we adopt a modified Point-MAE~\cite{pang2022PointMAE}, whose encoder produces a 384-dimensional global latent representation together with a geometric center. 
The autoencoder is pretrained solely for point cloud reconstruction and remains fixed during adversarial optimization to ensure a stable latent embedding space.

For each input point cloud, the sparse reconstruction coefficient $\alpha^\star$ is computed once via $\ell_1$-regularized optimization and kept fixed throughout the attack process, with the sparsity weight set to $\lambda_{\mathrm{spa}} = 0.1$. 
Adversarial perturbations are then optimized in the compact subspace by updating the low-rank parameters $U$ and $\Gamma$ using the Adam optimizer for 2{,}000 iterations with a learning rate of $1\text{e}{-2}$. 
The regularization weights are set to $\lambda_{\mathrm{rank}} = 1\text{e}{-4}$ and $\lambda_{\mathrm{ort}} = 1\text{e}{-3}$ to encourage compactness and stability of the perturbation subspace.

The perceptual loss is defined as the sum of the Chamfer distance (CD) and $0.1$ times the Hausdorff distance (HD), with the perceptual weight $\lambda_{\mathrm{per}} = 1$. 
After decoding the perturbed latent representation, the resulting adversarial point cloud is clipped to satisfy the prescribed $\ell_\infty$ perturbation budget with respect to the original input. 
All experiments are conducted on a workstation equipped with eight NVIDIA RTX~3090 GPUs.

\firstpara{Datasets}  
We evaluate CoSA on ModelNet40~\cite{wu-2015-mn40} and ScanObjectNN~\cite{uy2019revisiting}. 
ModelNet40 contains 9,843 training and 2,468 testing point clouds from 40 CAD object categories. 
In contrast, ScanObjectNN consists of 11,416 training and 2,882 testing samples collected from real-world indoor scans, featuring realistic noise, occlusion, and background clutter. 
Following~\cite{Xiang-2019-Generating}, all point clouds are uniformly resampled to 1,024 points.

\firstpara{Victim DNN Classifiers}
We consider four representative point cloud classifiers that cover diverse architectural paradigms: the MLP-based PointNet~\cite{Qi-2017-Pointnet}, the graph-based DGCNN~\cite{Wang-2019-DGCNN}, the transformer-based PCT~\cite{guo2021pct}, and the state-space model PointMamba~\cite{liang2024pointmamba}. 
All models are trained using the standard training protocols reported in their original papers to ensure fair and reproducible evaluation.

\begin{table*}[t]
\centering
\caption{
Attack success rates (ASR, \%) of different methods under four defense strategies on ModelNet40 and ScanObjectNN. 
Results include both direct (white-box) attacks and transfer attacks, where adversarial examples are generated on PointNet and evaluated on PointNet, DGCNN, PCT, and PointMamba, under two $\ell_\infty$ budgets ($\epsilon\!=\!0.18,0.45$). 
Bold indicates the highest performance.
}
\vspace{-1mm}
\label{tab:defense_asr_dual_budget}
\setlength{\tabcolsep}{1.2mm}{
\scalebox{1}{
\begin{tabular}{@{} c l|cccc|cccc|cccc|cccc @{}}
\toprule
\multirow{3}{*}{\makecell{Victim\\Network}} &
\multirow{3}{*}{\makecell{Attack\\Method}} &
\multicolumn{8}{c|}{\textbf{ModelNet40}} &
\multicolumn{8}{c}{\textbf{ScanObjectNN}} \\
\cmidrule(lr){3-10}\cmidrule(lr){11-18}
& & \multicolumn{4}{c|}{$\epsilon = 0.18$} & \multicolumn{4}{c|}{$\epsilon = 0.45$}
  & \multicolumn{4}{c|}{$\epsilon = 0.18$} & \multicolumn{4}{c}{$\epsilon = 0.45$} \\
\cmidrule(lr){3-6}\cmidrule(lr){7-10}\cmidrule(lr){11-14}\cmidrule(lr){15-18}
& & SRS & SOR & DUP\text{-}Net & AT
  & SRS & SOR & DUP\text{-}Net & AT
  & SRS & SOR & DUP\text{-}Net & AT
  & SRS & SOR & DUP\text{-}Net & AT \\
\midrule

\multirow{7}{*}{\makecell{PointNet \\ $\downarrow$ \\ PointNet}}
& 3D\text{-}Adv     & 26.6 & 17.5 & 12.6 & 5.0  & 1.6  & 17.2 & 12.4 & 1.0  & 32.8 & 24.5 & 27.7 & 15.1 & 34.1 & 27.2 & 38.4 & 18.6 \\
& KNN               & 35.7 & 18.6 & 27.8 & 15.3 & 36.2 & 19.9 & 29.7 & 19.3 & 79.9 & 73.0 & 63.7 & 40.6 & 79.8 & 72.7 & 64.1 & 43.4 \\
& AdvPC             & 60.5 & 36.8 & \textbf{40.1} & 22.0 & 63.2 & 33.7 & 38.9 & 25.3 & 71.4 & 63.4 & 52.5 & 36.8 & 72.1 & 63.5 & 52.7 & 35.8 \\
& AOF               & 63.9 & 33.5 & 35.7 & 19.6 & 64.5 & 32.3 & 34.1 & 22.2 & 78.8 & 71.5 & 58.7 & 41.1 & 79.5 & 70.2 & 56.7 & 38.2 \\
& PF\text{-}Attack  & 53.2 & 39.5 & 19.5 & 10.4 & 58.7 & 33.9 & 22.5 & 14.4 & 69.7 & 54.3 & 56.8 & 39.7 & 70.6 & 53.5 & 56.2 & 37.9 \\
& CFG               & 75.3 & \textbf{57.8} & 35.8 & 22.3 & 83.2 & \textbf{67.6} & 46.0 & 30.5 & 84.8 & 67.2 & 64.2 & 44.9 & 85.3 & 69.0 & 67.1 & 45.6 \\
& Ours              & \textbf{95.2} & 42.3 & 37.1 & \textbf{23.5} & \textbf{98.9} & 60.5 & \textbf{48.8} & \textbf{38.2} & \textbf{96.7} & \textbf{78.5} & \textbf{77.2} & \textbf{46.7} & \textbf{99.3} & \textbf{80.5} & \textbf{77.3} & \textbf{46.6} \\
\midrule

\multirow{7}{*}{\makecell{PointNet \\ $\downarrow$ \\ DGCNN}}
& 3D\text{-}Adv     & 11.9 & 11.5 & 12.7 & 7.0  & 11.9 & 11.4 & 11.8 & 6.7  & 23.1 & 24.6 & 24.7 & 14.8 & 23.0 & 24.2 & 25.5 & 14.4 \\
& KNN               & 29.6 & 17.9 & 19.9 & 10.9 & 29.3 & 18.2 & 21.1 & 13.8 & 53.4 & 48.2 & 40.3 & 31.2 & 56.6 & 50.4 & 43.2 & 22.6 \\
& AdvPC             & 23.7 & 16.4 & 19.5 & 10.7 & 24.5 & 16.3 & 19.5 & 12.4 & 43.4 & 40.7 & 38.4 & 29.8 & 44.1 & 40.8 & 39.7 & 21.4 \\
& AOF               & 46.1 & 36.1 & 35.6 & 19.6 & 47.2 & 36.0 & 38.2 & 24.6 & 49.1 & 50.5 & 42.9 & 32.5 & 51.7 & 51.1 & 47.3 & 24.7 \\
& PF\text{-}Attack  & 25.1 & 25.5 & 21.7 & 11.3 & 30.6 & 23.4 & 25.1 & 16.3 & 41.1 & 40.2 & 36.3 & 23.3 & 41.3 & 40.9 & 38.6 & 22.2 \\
& CFG               & 44.2 & 41.1 & 42.0 & 23.1 & 54.3 & 51.7 & 45.6 & \textbf{36.1} & 50.4 & 48.5 & 47.1 & 35.1 & 53.1 & 50.1 & 48.4 & 36.6 \\
& Ours              & \textbf{58.2} & \textbf{55.2} & \textbf{54.7} & \textbf{30.1} & \textbf{59.8} & \textbf{57.9} & \textbf{49.0} & 34.2 & \textbf{68.6} & \textbf{57.9} & \textbf{56.4} & \textbf{36.7} & \textbf{69.8} & \textbf{65.4} & \textbf{57.6} & \textbf{37.8} \\
\midrule

\multirow{7}{*}{\makecell{PointNet \\ $\downarrow$ \\ PCT}}
& 3D\text{-}Adv     & 19.2 & 18.0 & 17.8 & 9.8  & 18.8 & 18.3 & 18.1 & 11.7 & 20.7 & 20.8 & 24.8 & 14.6 & 21.5 & 21.6 & 25.3 & 15.1 \\
& KNN               & 34.2 & 22.5 & 22.2 & 13.1 & 34.2 & 23.1 & 21.6 & 14.0 & 58.5 & 47.7 & 41.6 & 27.0 & 58.8 & 48.7 & 32.4 & 25.6 \\
& AdvPC             & 28.0 & 21.5 & 21.3 & 11.7 & 28.9 & 22.4 & 21.0 & 13.6 & 52.1 & 49.3 & 39.7 & 25.8 & 54.3 & 49.6 & 38.1 & 24.4 \\
& AOF               & 38.9 & 31.7 & 31.3 & 17.2 & 41.6 & 32.3 & 32.7 & 21.3 & 57.2 & 56.5 & 43.2 & 29.2 & 57.6 & 56.8 & 46.7 & 32.3 \\
& PF\text{-}Attack  & 28.0 & 24.2 & 23.4 & 12.9 & 31.0 & 20.4 & 24.6 & 16.4 & 49.3 & 46.9 & 46.4 & 25.3 & 50.2 & 46.8 & 42.7 & 34.8 \\
& CFG               & 34.0 & 33.1 & 36.2 & 19.1 & 52.1 & 43.9 & \textbf{48.1} & 33.9 & 55.5 & 50.1 & 48.2 & 29.7 & 56.1 & 52.5 & 46.5 & 28.6 \\
& Ours              & \textbf{53.2} & \textbf{35.1} & \textbf{36.5} & \textbf{20.1} & \textbf{58.7} & \textbf{50.3} & 46.4 & \textbf{36.7} & \textbf{67.2} & \textbf{58.7} & \textbf{52.3} & \textbf{36.4} & \textbf{64.1} & \textbf{62.3} & \textbf{55.3} & \textbf{36.5} \\
\midrule

\multirow{7}{*}{\makecell{PointNet \\ $\downarrow$ \\ PointMamba}}
& 3D\text{-}Adv     & 7.9  & 7.3  & 9.8  & 5.4  & 8.0  & 7.2  & 11.1 & 6.9  & 15.4 & 15.7 & 17.2 & 11.5 & 15.6 & 15.8 & 17.5 & 11.8 \\
& KNN               & 27.4 & 15.2 & 22.1 & 12.2 & 28.1 & 15.2 & 21.4 & 13.9 & 49.1 & 38.4 & 41.2 & 30.7 & 47.9 & 39.0 & 40.7 & 29.4 \\
& AdvPC             & 32.2 & 18.7 & 23.9 & 14.8 & 31.1 & 18.4 & 22.4 & 16.5 & 40.8 & 37.3 & 43.8 & 32.3 & 42.8 & 37.3 & 43.5 & 31.0 \\
& AOF               & 38.3 & 37.6 & 36.2 & 22.1 & 38.6 & 40.6 & 35.8 & 26.3 & 49.5 & 47.5 & 46.6 & 35.1 & 48.4 & 47.2 & 46.5 & 33.9 \\
& PF\text{-}Attack  & 42.8 & 39.4 & 39.8 & 21.9 & 39.8 & 37.7 & 35.6 & 23.1 & 43.7 & 38.2 & 40.9 & 27.8 & 45.0 & 38.9 & 41.9 & 27.1 \\
& CFG               & 54.7 & 51.8 & 49.6 & 27.3 & 62.3 & 57.5 & 53.4 & 35.8 & 45.3 & 39.3 & 41.1 & 29.7 & 49.5 & 43.7 & 43.9 & 29.6 \\
& Ours              & \textbf{65.8} & \textbf{55.9} & \textbf{53.2} & \textbf{29.2} & \textbf{67.2} & \textbf{62.1} & \textbf{53.5} & \textbf{38.4} & \textbf{65.2} & \textbf{61.2} & \textbf{62.7} & \textbf{40.3} & \textbf{65.8} & \textbf{60.7} & \textbf{60.9} & \textbf{39.7} \\
\bottomrule
\end{tabular}
}}
\end{table*}

\firstpara{Baselines}
We compare CoSA with eight state-of-the-art transferable point cloud attack methods: 3D-Adv~\cite{Xiang-2019-Generating}, KNN~\cite{tsai-2020-robust(smooth)}, AdvPC~\cite{hamdi2020advpc}, AOF~\cite{liu2022boosting}, PF-Attack~\cite{He-2023-PF}, ShapeAdv~\cite{lee-2020-Shapeadv}, MAT-Adv~\cite{tang2025mat}, and CFG~\cite{Pang2025-CFG}.

\firstpara{Evaluation Setting and Metrics}
We evaluate transferability using the attack success rate (ASR), defined as the proportion of adversarial point clouds that cause misclassification on a target model.
Unless otherwise specified, adversarial examples are generated on a surrogate model and then evaluated on unseen target architectures in a black-box manner.
We report results under two $\ell_\infty$ perturbation budgets, $\epsilon=0.18$ and $0.45$.

\subsection{Comparison with State-of-the-art Methods}

\firstpara{Performance on Transferability}
Tab.~\ref{tab:transfer_table_all} reports transfer attack results on ModelNet40 and ScanObjectNN under two $\ell_\infty$ budgets.
All methods achieve high success rates in white-box settings, confirming effective optimization on their respective surrogate models.
In black-box transfer scenarios, clear performance gaps emerge.
Across most source--target pairs, CoSA consistently attains the highest attack success rates and generally outperforms CFG, with improvements typically around 8 to 12 percent and reaching about 15 percent on more challenging target architectures such as PCT and PointMamba.
These gains are observed on both datasets and under both perturbation budgets, indicating that the improvement is not tied to a specific model or dataset.
Overall, the results demonstrate that constructing perturbations within a two-level compact subspace substantially enhances cross-architecture transferability.

\firstpara{Performance on Attack under Defense}
Tab.~\ref{tab:defense_asr_dual_budget} reports attack performance under four defense strategies: simple random sampling (SRS), statistical outlier removal (SOR), the denoiser and upsampler network (DUP-Net), and adversarial training (AT), using PointNet as both the source and target model.
As expected, all defenses reduce attack success rates to varying degrees.
Under SRS, DUP-Net, and AT, CoSA achieves the highest success rates among all compared methods, while CFG performs slightly better under SOR.
Notably, under denoising-based defenses such as SRS and DUP-Net, CoSA maintains a clear advantage over other transfer-oriented attacks.
These results indicate that compact subspace perturbations are less susceptible to being removed by defense operations, leading to improved robustness under common defenses.

\firstpara{Performance on Transferability under Defense}
Building on the white-box defense results above, the lower part of Tab.~\ref{tab:defense_asr_dual_budget} summarizes transferability under defense, where adversarial examples generated on PointNet are evaluated on unseen target architectures including DGCNN, PCT, and PointMamba.
Across most defenses and perturbation budgets, CoSA consistently outperforms CFG and AOF in transfer settings.
The performance gains are particularly evident under denoising-based defenses, where CoSA often improves ASR by around 10 percent and reaches larger margins in several source--target combinations.
These results suggest that restricting perturbations to a compact semantic subspace enhances perturbation stability after defense filtering and supports robust cross-architecture transfer even in defended scenarios.

\begin{figure*}[!t] 
\centering
\includegraphics[width=1\linewidth]{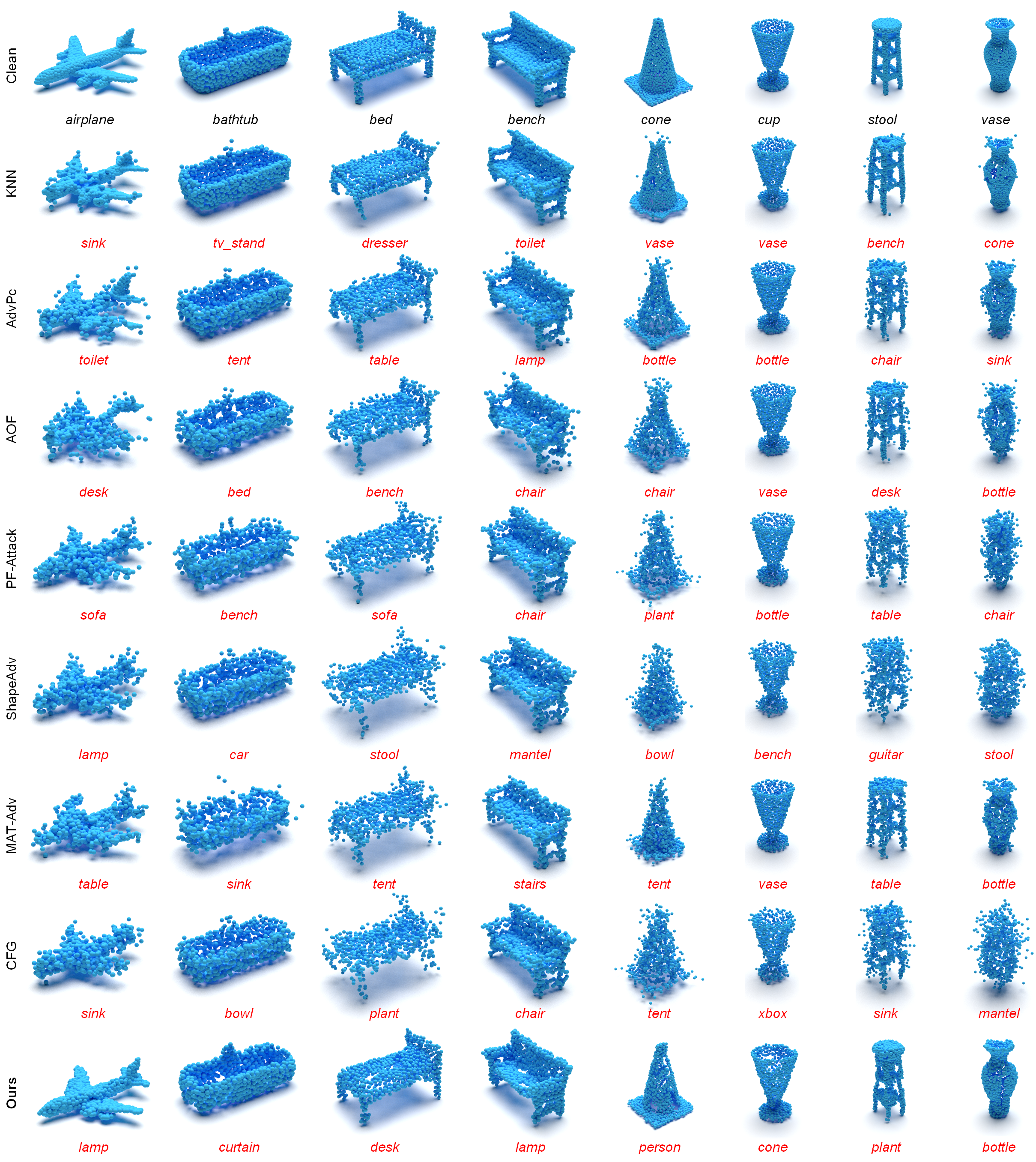}
\caption{
Visualization of original and adversarial point clouds generated by various attack methods targeting PointNet on ModelNet40. Ground-truth and predicted labels are shown below each example in black and red, respectively.
}
\label{fig:CoSA_Adv_examples} 
\end{figure*}

\firstpara{Visualization}
Fig.~\ref{fig:CoSA_Adv_examples} presents qualitative comparisons of adversarial point clouds generated by different attack methods on several representative object categories.
Consistent with the quantitative results in Tab.~\ref{tab:transfer_table_all}, CoSA achieves strong transferability while maintaining visually coherent shapes.
Compared with other methods, CoSA produces adversarial examples with fewer isolated points or irregular distortions, and the overall geometric structure of the objects remains largely intact.
Notably, no obvious outliers or unnatural artifacts are observed, even in regions with fine geometric details.
These observations indicate that constraining adversarial modifications to a compact subspace helps preserve intrinsic shape structure, which in turn supports effective cross-model transferability.

\subsection{Ablation Studies and Other Analysis}

\firstpara{Importance of Two Subspaces}
Tab.~\ref{tab:subspace_transfer} studies how the base subspace $\mathcal{B}$ and the perturbation subspace $\mathcal{S}$ contribute to cross-model transferability on ModelNet40.
When both $\mathcal{B}$ and $\mathcal{S}$ are removed, CoSA reduces to an unconstrained latent attack and exhibits the weakest black-box performance, with transfer ASR staying in the low range on DGCNN, PCT, and PointMamba under both budgets.
Introducing only $\mathcal{S}$ provides a modest gain, indicating that a structured perturbation parameterization alone helps but is insufficient to align perturbations with model-agnostic semantics.
In contrast, enabling only $\mathcal{B}$ leads to a substantial improvement across all target architectures, suggesting that prototype-guided semantic anchoring is the primary factor for transfer.
Finally, using both subspaces consistently achieves the best results at $\epsilon\!=\!0.18$ and $0.45$, showing that $\mathcal{B}$ offers semantic regularity while $\mathcal{S}$ further stabilizes the perturbation directions, and their combination yields the most transferable adversarial examples.

\begin{figure*}[!t] 
\centering
\includegraphics[width=0.78\textwidth]{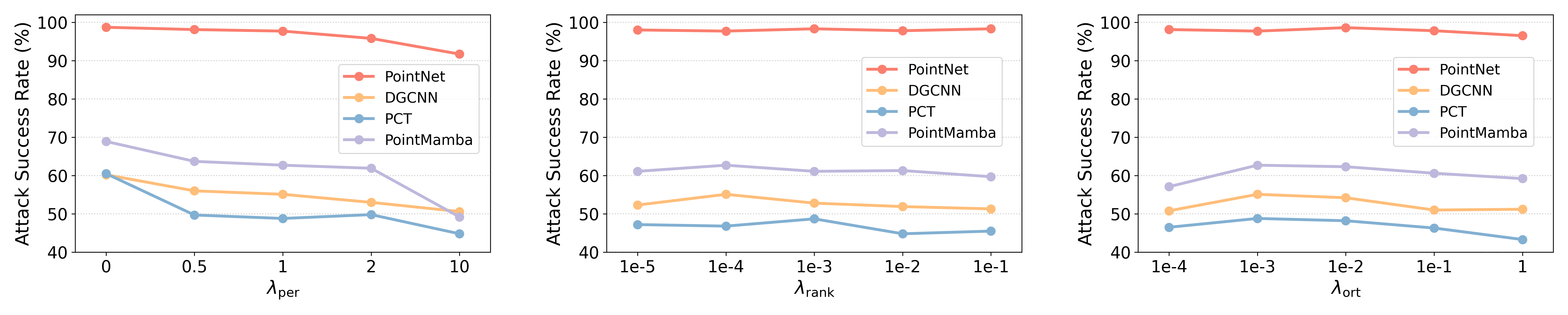}

\caption{
Effect of the perceptual weight $\lambda_{\mathrm{per}}$, the rank regularization weight $\lambda_{\mathrm{rank}}$, and the orthogonality weight $\lambda_{\mathrm{ort}}$ on the ASR of CoSA under $\epsilon = 0.18$.
The plot reports white-box ASR on PointNet and transfer ASR on DGCNN, PCT, and PointMamba.
}
\label{fig:lambdas_lines} 
\end{figure*}

\firstpara{Parameter Analysis}
We further analyze the sensitivity of CoSA to its key hyperparameters, including the perceptual weight $\lambda_{\mathrm{per}}$, the rank regularization weight $\lambda_{\mathrm{rank}}$, and the orthogonality weight $\lambda_{\mathrm{ort}}$, as summarized in Fig.~\ref{fig:lambdas_lines}.
Overall, CoSA demonstrates stable behavior across a broad range of values for all three parameters, indicating that the proposed framework does not rely on delicate tuning.

As $\lambda_{\mathrm{per}}$ increases, transferability gradually decreases on black-box models, while white-box performance remains relatively stable.
This trend reflects the expected trade-off between geometric fidelity and perturbation freedom: stronger perceptual constraints limit the extent of admissible deformations and thus reduce the space of transferable perturbations.
In contrast, varying $\lambda_{\mathrm{rank}}$ or $\lambda_{\mathrm{ort}}$ leads to only mild changes in ASR across different target architectures.
This suggests that once a reasonable low-rank structure and approximate orthogonality are enforced, the optimization remains effective and does not critically depend on precise weighting.
Overall, these results indicate that CoSA maintains consistent transfer performance within a wide and practical range of hyperparameter settings.

\newcommand{\colw}{6mm} 
\begin{table}[t]
\centering
\begin{minipage}{0.47\textwidth}
\centering
\caption{
Ablation on the base subspace $\mathcal{B}$ and  perturbation subspace $\mathcal{S}$.
Performance is measured using ASR  on PointNet and its transfer to other models under two $\ell_\infty$ budgets  on ModelNet40.
}
\vspace{-2.5mm}
\label{tab:subspace_transfer}
\setlength{\tabcolsep}{1.1pt}
\scalebox{0.90}{
\begin{tabular}{@{} c c|cccc|cccc @{}}
\toprule
\multicolumn{2}{c|}{} &
\multicolumn{4}{c|}{$\epsilon = 0.18$} &
\multicolumn{4}{c}{$\epsilon = 0.45$} \\
\cmidrule(lr){1-2}\cmidrule(lr){3-6}\cmidrule(lr){7-10}
\multicolumn{1}{c}{\makebox[\colw][c]{$\mathcal{B}$}} &
\multicolumn{1}{c|}{\makebox[\colw][c]{$\mathcal{S}$}} &
\multicolumn{1}{c}{PointNet} &
\multicolumn{1}{c}{DGCNN} &
\multicolumn{1}{c}{PCT} &
\multicolumn{1}{c|}{\makecell{Point-\\Mamba}} &
\multicolumn{1}{c}{PointNet} &
\multicolumn{1}{c}{DGCNN} &
\multicolumn{1}{c}{PCT} &
\multicolumn{1}{c}{\makecell{Point-\\Mamba}} \\
\midrule
\makebox[\colw][c]{\ding{55}} & \makebox[\colw][c]{\ding{55}} 
& \textcolor{gray}{92.7} & 27.4 & 18.5 & 23.6 
& \textcolor{gray}{95.1} & 30.5 & 19.7 & 25.8 \\
\makebox[\colw][c]{\ding{55}} & \makebox[\colw][c]{\checkmark} 
& \textcolor{gray}{96.1} & 32.2 & 23.8 & 25.2 
& \textcolor{gray}{100.0} & 35.7 & 26.1 & 32.3 \\
\makebox[\colw][c]{\checkmark} & \makebox[\colw][c]{\ding{55}} 
&\textcolor{gray}{95.3} & 52.3 & 45.9 & 59.4 
&\textcolor{gray}{100.0} & 56.1 & 51.4 & 61.7 \\

\makebox[\colw][c]{\checkmark} & \makebox[\colw][c]{\checkmark} &
\textcolor{gray}{97.7} & \textbf{55.1} & \textbf{48.8} & \textbf{62.7} &
\textcolor{gray}{100.0} & \textbf{58.1} & \textbf{54.2} & \textbf{65.6} \\
\bottomrule
\end{tabular}
}
\end{minipage}
\vspace{-1mm}
\end{table}

\begin{figure}[!t] 
\centering
\includegraphics[width=1.0\linewidth]{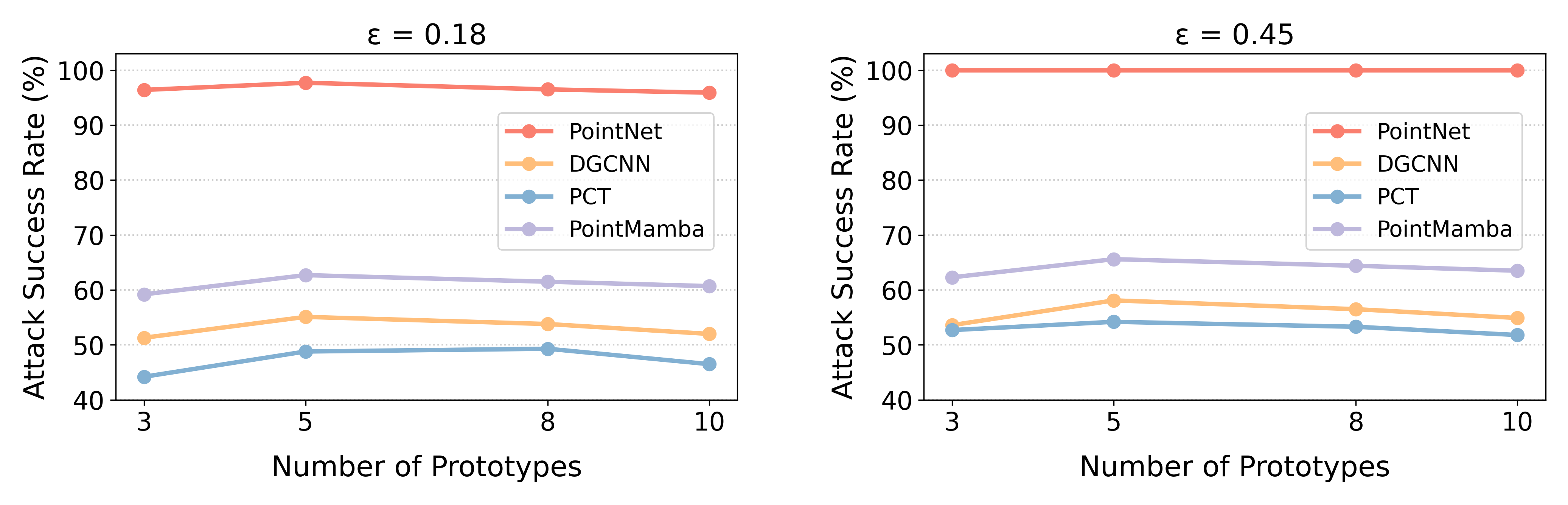}
\caption{
Parameter analysis of prototype number $m_y$ in the base subspace.
Performance is measured using ASR on PointNet and its transfer to other models under two $\ell_\infty$ budgets on ModelNet40.
}
\label{fig:proto_k_lines} 
\end{figure}

\firstpara{Effect of Prototype Number $m_y$}
Fig.~\ref{fig:proto_k_lines} reports the attack success rate under two $\ell_\infty$ budgets as the number of prototypes $m_y$ varies.
Increasing $m_y$ from 3 to 5 consistently improves both white-box and transfer performance, suggesting that a moderate number of prototypes helps capture richer intra-class semantic variations.
When $m_y$ becomes larger, the performance gains saturate or slightly decline, which can be attributed to reduced compactness of the base subspace and the introduction of redundant or less representative directions.
These observations indicate that an overly large prototype set does not further benefit transferability and may weaken the semantic coherence of the subspace.
Based on this trade-off between representation diversity and compactness, we adopt $m_y=5$ as the default setting.

\begin{figure}[!t] 
\centering
\includegraphics[width=1.0\linewidth]{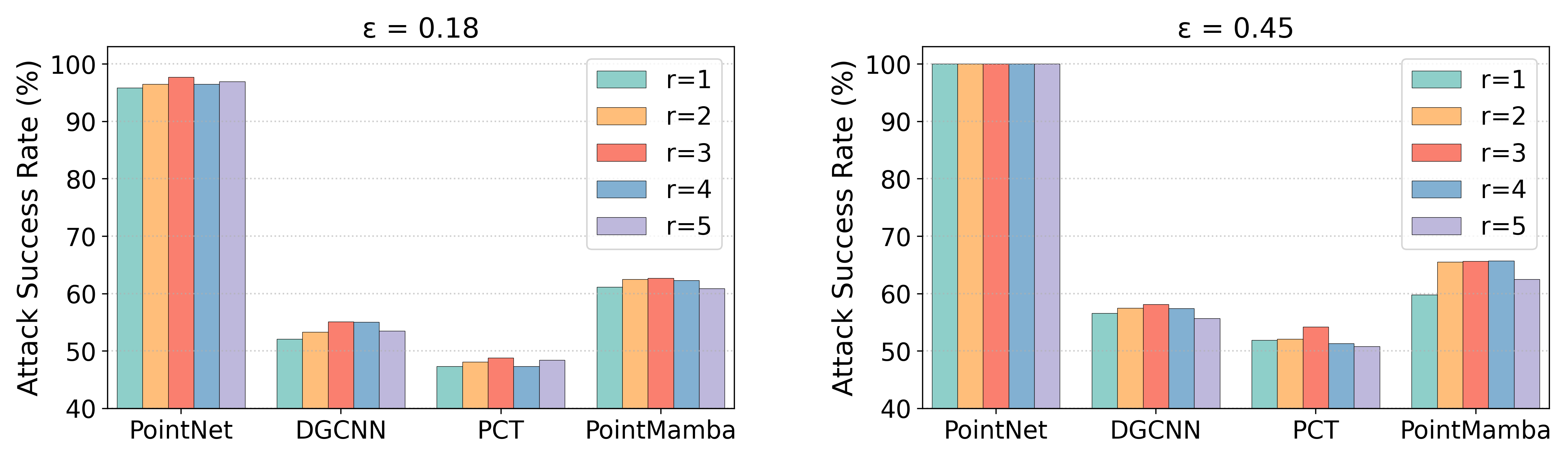}
\caption{
Parameter analysis of the rank $r$ in the perturbation subspace.
Performance is measured using ASR on PointNet and its transfer to other models under two $\ell_\infty$ budgets  on ModelNet40.
}
\label{fig:lowrank} 
\end{figure}

\firstpara{Effect of Low-rank Parameter $r$}
Fig.~\ref{fig:lowrank} examines the influence of the low-rank parameter $r$ on attack success.
As $r$ increases from 1 to 3, the ASR consistently improves across target models, indicating that a small number of coordinated perturbation directions enhances flexibility and transferability.
Further increasing $r$ yields diminishing returns and may slightly degrade performance, likely because additional directions introduce redundancy and reduce the structural regularity of the perturbation subspace.
These results suggest that effective transfer does not require a high-rank perturbation space, but rather a compact set of well-aligned directions.
We therefore set $r=3$ as the default to balance perturbation expressiveness and subspace regularity.

\begin{figure}[!t] 
\centering
\includegraphics[width=1\linewidth]{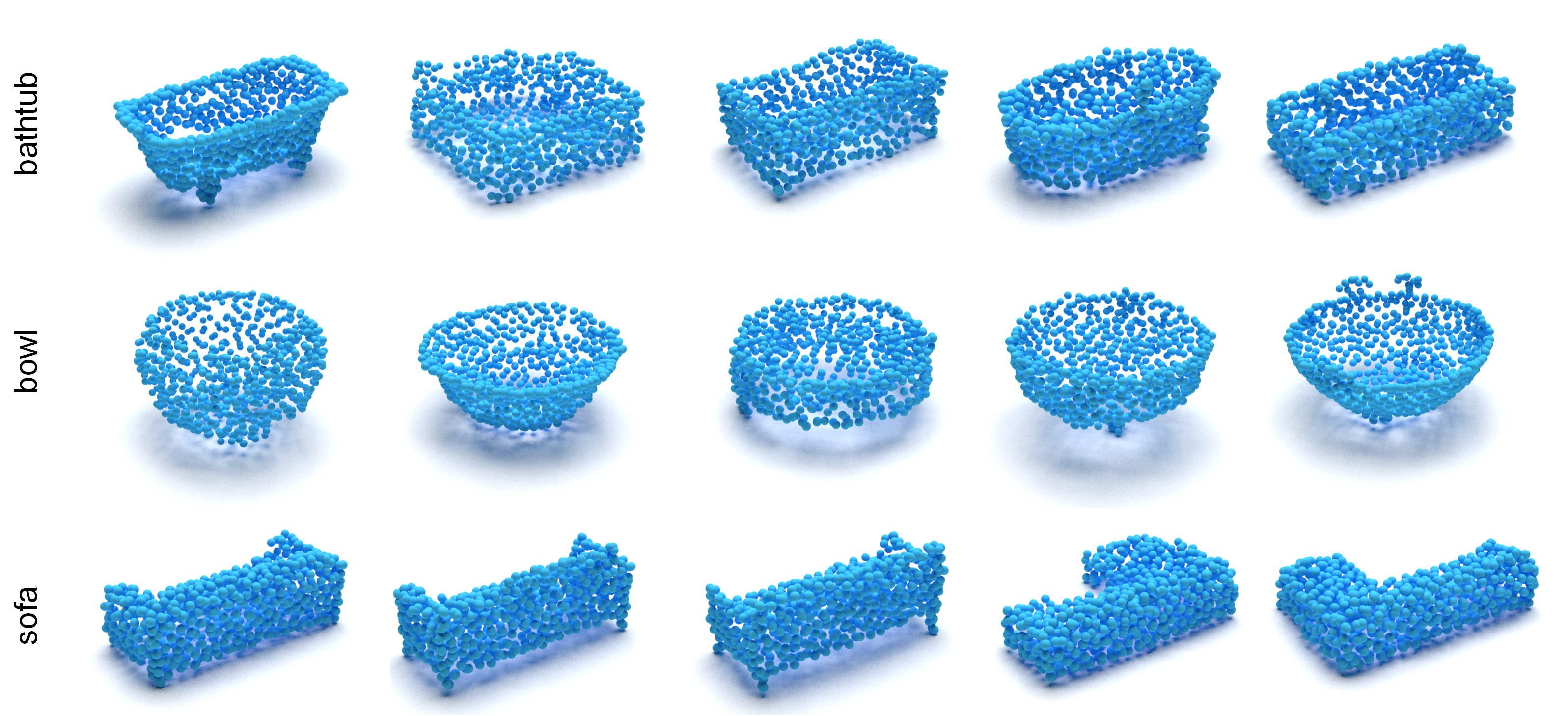}
\caption{Visualization of prototypes from three representative categories, each containing five decoded prototypes.}
\label{fig:Proto_vis} 
\end{figure}

\firstpara{Visualization of Prototypes}
To qualitatively examine the learned base subspace, Fig.~\ref{fig:Proto_vis} visualizes several class-wise prototypes.
The prototypes exhibit meaningful structural variations within each category, capturing different yet semantically consistent geometric patterns.
Importantly, these variations remain compact and do not introduce implausible or distorted shapes.
Such prototype diversity provides a stable semantic anchor for adversarial perturbations, allowing modifications to generalize across samples with similar geometric structures.
This qualitative evidence supports the role of the prototype-guided base subspace in facilitating transferable adversarial behavior.

\firstpara{Overhead}
Tab.~\ref{tab:time_comparison} reports the average time required to generate an adversarial point cloud when attacking PointNet.
CoSA incurs computational overhead comparable to existing transferable attack methods.
The additional cost mainly stems from latent-space optimization and low-rank parameter updates, while no expensive per-iteration operations are introduced.
Overall, the runtime remains practical for large-scale evaluation, demonstrating that CoSA achieves improved transferability without sacrificing computational efficiency.

\firstpara{Imperceptibility Performance}
Tab.~\ref{tab:cdhd_modelnet40} reports the geometric distortions of different attacks when generating adversarial point clouds on PointNet, measured by CD, HD, and $\ell_2$ (lower is better). Since 3D-Adv is not designed for transferability, it naturally achieves the lowest distortions across all metrics. Among transfer-oriented attacks, a clear tradeoff emerges: methods that pursue stronger transferability often incur larger geometric distortions. Within this group, CoSA yields distortions close to AOF, and is consistently lower than CFG and PF-Attack on all three metrics, indicating improved imperceptibility under comparable transfer-oriented settings. Although CoSA cannot match the extremely low distortion of 3D-Adv, it offers a more favorable balance by maintaining competitive imperceptibility while achieving significantly stronger transferability.

\firstpara{Effect of AE}
We further evaluate the impact of the AE backbone in CoSA by replacing the modified Point-MAE with PointFlow.
As shown in Tab.~\ref{tab:ae_asr_recon_compact}, PointFlow attains slightly lower reconstruction fidelity, while achieving comparable transfer ASR across target models.
This result suggests that CoSA is not tightly coupled to a particular AE design.
Instead, its transferability primarily comes from the proposed compact subspace formulation, which provides a stable semantic space for constructing transferable perturbations.

\begin{table}[t] 
\centering 
\begin{minipage}{0.47\textwidth} 
\centering 
\caption{The average time required by different methods to generate an adversarial example to attack PointNet on ModelNet40.}
\vspace{-3mm}
\label{tab:time_comparison} 
\setlength{\tabcolsep}{4pt} 
\scalebox{1.0}{ 
\begin{tabular}{ccccccc} 
\toprule 
Attack Method & 3D-Adv & KNN & PF-Attack &  CFG & Ours \\ 
\midrule 
Time (second) & 12.8 & 9.6 & 18.0 & 6.2 & 10.3  \\ 
\bottomrule 
\end{tabular} 
} 
\end{minipage} 
\vspace{-2mm}
\end{table}


\begin{table}[!t]
\centering
\centering
\caption{
Imperceptibility comparison of different attack methods on ModelNet40 using PointNet as the white-box source model.
We report the average geometric distortions (CD, HD) and  $\ell_2$ distances for adversarial examples.
}
\vspace{-2mm}
\label{tab:cdhd_modelnet40}
\setlength{\tabcolsep}{4pt}
\scalebox{1.0}{%
\begin{tabular}{lccc}
\toprule
Method      & CD ($\times 10^{-3}$) & HD ($\times 10^{-2}$) & $\ell_2$ ($\times 10^{-2}$) \\
\midrule
3D-Adv          & 0.041 & 0.609 & 0.272 \\
KNN             & 0.461 & 2.132 & 1.420 \\
AdvPC           & 0.889 & 0.710 & 2.381 \\
AOF             & 1.661 & 1.251 & 4.215 \\
PF-Attack       & 2.529 & 2.004 & 6.972 \\
CFG             & 3.561 & 2.611 & 9.307 \\
\textbf{Ours}   & 1.749 & 1.074 & 6.816 \\
\bottomrule
\end{tabular}%
}
\end{table}

\begin{table}[!t]
\centering
\begin{minipage}{0.47\textwidth}
\centering
\caption{
Comparison of AE backbones in CoSA. For each backbone, we report the average transfer ASR ($\epsilon{=}0.18$, targets: DGCNN, PCT, PointMamba) and the reconstruction quality of the autoencoder, measured by CD and HD, on ModelNet40.
}
\vspace{-3mm}
\label{tab:ae_asr_recon_compact}
\setlength{\tabcolsep}{6pt}
\scalebox{0.98}{
\begin{tabular}{cccc}
\toprule
  AE & Avg. ASR (\%) $\uparrow$ & CD ($\times10^{-4}$) $\downarrow$& HD ($\times10^{-3}$)$\downarrow$ \\
\midrule
PointFlow & 53.1 & 8.5 & 3.8 \\
Point-MAE  & 55.5 & 6.6 & 4.8 \\
\bottomrule
\end{tabular}
}
\end{minipage}
\end{table}

%% file: 6_conclusion.tex
\section{Conclusion}
In this paper, we have presented CoSA, a transferable adversarial attack framework for point clouds.
By constraining perturbations to a compact subspace that captures shared, model-agnostic variations, CoSA improves transferability across diverse neural architectures.
Extensive experiments demonstrate that CoSA achieves strong cross-model transferability and outperforms state-of-the-art point cloud attack methods.
Future work will extend this compact subspace formulation to physical-world attacks and robust point cloud defenses.